\documentclass[10pt,twocolumn,letterpaper]{article}

\usepackage{cvpr}
\usepackage{times}
\usepackage{epsfig}
\usepackage{graphicx}
\usepackage{amsmath}
\usepackage{amssymb}
\usepackage{subfig}
\usepackage{booktabs}
\usepackage{bm}
\usepackage{color}
\usepackage{color}
\usepackage{xcolor}
\usepackage{algorithmic,algorithm}
\usepackage{booktabs}
\usepackage{threeparttable}
\usepackage{multirow}

\usepackage[pagebackref=true,breaklinks=true,letterpaper=true,colorlinks,bookmarks=false]{hyperref}

\cvprfinalcopy 

\def\cvprPaperID{****} 

\ifcvprfinal\fi
\begin{document}

\title{Combining Background Subtraction Algorithms with Convolutional Neural Network}


\author{Dongdong Zeng,~Ming Zhu                         and~Arjan Kuijper  \\
{\tt\small zengdongdong13@mails.ucas.edu.cn, zhu\_mingca@163.com, arjan.kuijper@igd.fraunhofer.de}
}

\maketitle


\begin{abstract}
Accurate and fast extraction of foreground object is a key prerequisite for a wide range of computer vision applications such as object tracking and recognition. Thus, enormous background subtraction methods for foreground object detection have been proposed in recent decades.  However, it is still regarded as a tough problem due to a variety of challenges such as illumination variations, camera jitter,  dynamic backgrounds, shadows, and so on. Currently, there is no single method that can handle all the challenges in a robust way.
In this letter,  we try to solve this problem from a new perspective by combining  different state-of-the-art background subtraction algorithms to create a more robust and more advanced foreground detection algorithm. More specifically,  an encoder-decoder fully convolutional neural network architecture is trained to automatically learn how to leverage the characteristics of different algorithms  to fuse the results produced by  different  background subtraction algorithms  and  output a more precise result.
Comprehensive experiments evaluated on the CDnet 2014 dataset demonstrate  that the proposed method outperforms all the considered single background subtraction algorithm.
And we  show that our solution is more  efficient  than other combination strategies.

\end{abstract}

\section{Introduction}

Foreground object detection for a stationary camera is one of the essential tasks in  many computer vision and video analysis applications such as object tracking, activity recognition, and human-computer interactions.
As the first step in  these high-level operations, the accurate extraction of foreground object directly affects the subsequent operations.
Background subtraction (BGS) is the most  popular technology used for foreground object detection. The performance of foreground extract highly depended on the reliability of background modeling. In the past decades, various background modeling strategies have been proposed by researchers \cite{survey1, survey2}.
One of the most commonly used assumptions is that the  probability density function of a pixel intensity is a Gaussian or Mixture of Gaussians (MOG), as proposed in \cite{SingleGaussian, GMM}.
A non-parametric  approach using Kernel Density Estimation (KDE) technique was proposed in \cite{KDE}, which estimates the probability density function at each pixel from many samples without any prior assumptions.
The codebook-based method was introduced by Kim \textit{et al.} \cite{codebook},  where the background  pixel value is modeled into codebooks which represent a compressed form of background model in a long image sequence time.
The more recent ViBe algorithm \cite{VIBE}  was built on the similar principles as the GMMs,  the authors try to store the distribution with a random  collection of samples.
If the pixel in the new input frame matches with a  proportion of its background samples, it is considered  to be background and may be selected for model updating.
St-Charles \textit{et al.} \cite{subsense}  improved the method by using Local Binary Similarity Patterns  features and color features, and   a pixel-level feedback loop is used to  adjust the internal parameters. Recently, some deep learning based methods was proposed \cite{CNNBGS1, DeepBS, myMFCN, FgSegNet}, which show  state-of-the-art performance, however, these methods need the ground truth constructed by human to train the model, so, it could be argued whether they are useful in the practical applications.

\begin{figure*}[htbp]
\centering
\includegraphics[width=1.0\linewidth]{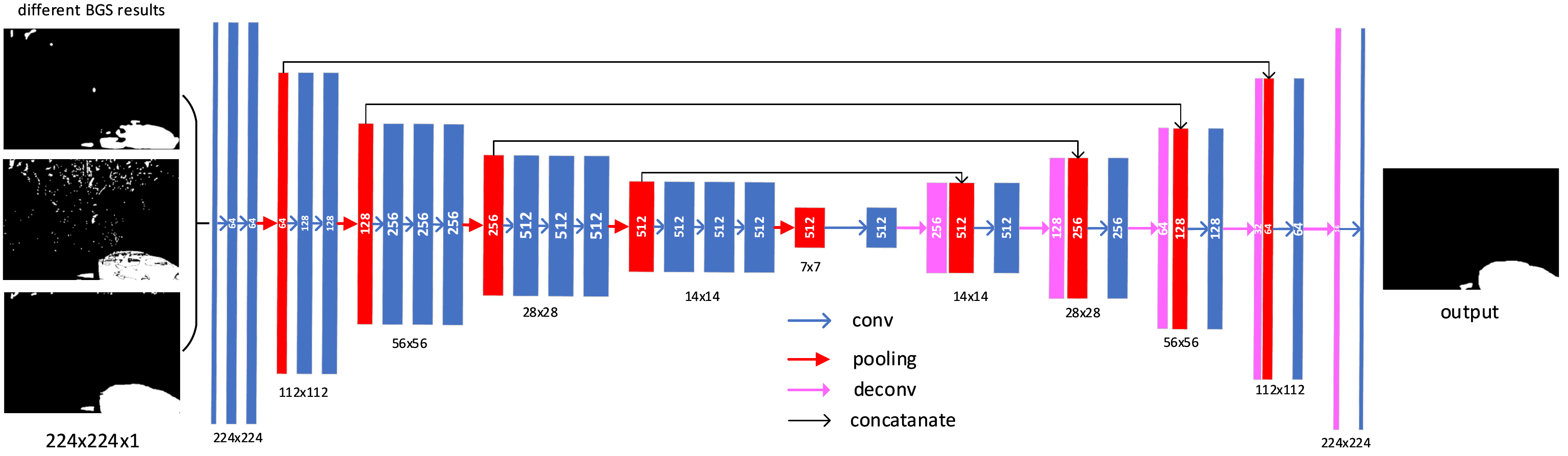}  
\caption{Proposed encoder-decoder fully convolutional neural network for combining different background subtraction results. The inputs are  foreground/backgroud masks generated by SuBSENSE \cite{subsense}, FTSG \cite{FTSG}, and CwisarDH \cite{cwisardH} algorithms, respectively. The encoder is a VGG16 \cite{VGG} network without fully connected layers. The max pooling operations separate it into five stages. To  effectively utilize  different levels of  feature information from different stages, a set of concatanate and deconvolution operations is used to aggregate different scale features, so that more category-level information and fine-grain details are represented.
}
\label{iutis3}
\end{figure*}

Despite the numerous BGS methods that have been proposed,  there is no single algorithm can deal with all these challenges in the real-world scenario.  Recently, some works try to combine different BGS algorithms to get better performance. They  fuse the output results from different algorithms with some strategies  to produce a more accurate foreground segmentation result. For example, in \cite{CDnet2014}, a pixel-based majority vote (MV) strategy is used to combine the results from different algorithms. They showed that except for some special algorithms, the majority vote result outperforms  every single method.   In \cite{bianco2017combination}, a fusion strategy called IUTIS based on genetic programming (GP) is proposed.  During the learning stage, a set of unary, binary, and $n$-ary functions is embedded into the GP framework to determine the combination (\textit{e.g.} logical AND, logical OR) and post-processing (\textit{e.g.} filter operators)  operations performed on the  foreground/background masks generated by  different algorithms. It has been   shown that this solution outperforms all state-of-the-art BGS algorithms at present.

In the past few years, deep learning has revolutionized the field of computer vision. Deep convolutional neural networks (CNNs) were initially designed for the image classification task  \cite{CNN, he2016deep}. However, due to its powerful ability of extracting high-level feature representations, CNNs have been  successfully applied to other computer vision tasks such as  semantic segmentation \cite{SegSegnet, peng2017large}, saliency detection \cite{zhao2015saliency}, object tracking \cite{CNNTracking}, and so on. Inspired by this,
in this letter, we propose  an encoder-decoder fully convolutional neural network architecture (Fig. \ref{iutis3}) to combine the output results from different BGS algorithms. We show that the network can automatically learn to leverage the characteristics of different algorithms to produce a more accurate foreground/background mask.
To the best of our knowledge, this is the first attempt to apply CNNs to combine BGS algorithms.
Using the CDnet 2014 dataset \cite{CDnet2014}, we evaluate our method against numerous surveillance scenarios. The  experimental results show that the proposed method outperforms all the state-of-the-art BGS methods, and is superior to other combination strategies.

The rest of this letter is organized as follows. Section \ref{proposed_method_framework} reports the proposed method for BGS algorithms combination.
Section \ref{Experimental results} shows the experimental results carried out on the CDnet 2014 dataset. Conclusions follow in Section \ref{CONCLUSION}.

\section{PROPOSED METHOD}\label{proposed_method_framework}
In this section, we  give a detailed description of the proposed  encoder-decoder fully convolutional neural network architecture for  combining  BGS algorithms results.

\subsection{Network Architecture}
As shown in Fig. \ref{iutis3}, the proposed network is an U-net \cite{SegUnet} type architecture with an encoder network and a corresponding decoder network.
Be different from the original U-net, here, we use the VGG16 \cite{VGG} that trained on the Imagenet \cite{ImageNet} dataset as the encoder because some researches \cite{SegSegnet, iglovikov2018ternausnet} show that initializing the network with the weights trained on a large dataset   shows better performance than   trained from scratch with a randomly initialized weights.
The encoder  network VGG16 contains 13 convolutional layers coupled with ReLU activation functions. We  remove the  fully connected layers  in favor of retaining more spatial details  and reducing the network parameters.
The use of max pooling operations separates the VGG16 into five stages, feature maps of the same stage are generated by convolutions with $3\times3$ kernels,  the sizes and the number of channels  are shown in Fig. \ref{iutis3}. 

The main task of the decoder is to upsample the feature maps from the encoder to match with the input size. In contrast to  \cite{Segdeconvolution, SegSegnet}, who  use unpooling operation for upsampling, we use the deconvolution (transposed convolution) with stride 2 to double the size of a feature map.  The main advantage of deconvolution is that it does not  need to remember the pooling indexes from the encoder, thus reducing memory and computation requirements.
We know that CNNs provide multiple levels of abstraction in the feature hierarchies \cite{CNN, ma2015hierarchical}, the feature maps in the lower layers retain higher spatial resolution but only perceive low-level visual information like corners and edges, while the deeper layers can capture more high-level semantic information (object level or category level) but with less  fine-grained spatial details.
Advanced semantic features help to identify categories of image regions, while low-level visual features help to generate  detailed boundaries for accurate prediction.
To effectively use  multiple levels of feature information from different stages, in each upsampling procedure, the output of a deconvolution from the previous stage is concatenated with the corresponding  feature map in the encoder first. Then a convolution operation is applied on the concatenated features to make the channels the same with the encoder.
Finally,  transposed convolution is used to generate upsampled feature maps. This procedure is repeated 5 times. The final score map is then fed to a soft-max classifier to produce the foreground/background probabilities for each pixel.

For training the network, we use the class-balancing cross entropy loss function which was originally proposed in \cite{holistically} for contour detection.
Let's denote the training dataset as $S = \{ (X_n, Y_n) , n = 1, \ldots, N   \}$, where $X_n$ is the input sample, and $Y_n = \{ y_p^{(n)}  \in \{0,1\},  p = 1, \ldots, | X_n | \}$  is the corresponding labels.
Then the loss function is defined as follows:
\begin{equation}
\begin{aligned}
\mathcal{L}(\bm{W}) = - \beta\sum_{p \in Y_+}logPr(y_p=1|X;\bm{W}) \\
  -  ( 1 - \beta) \sum_{p \in Y_{-}}   logPr(y_p=0|X;\bm{W} ),
\end{aligned}
\end{equation}
where  $\beta = |Y_{-}|/|Y|$ and $1 - \beta = |Y_{+}| / |Y|  $. $Y_{+}$ and $Y_{-}$ denote the foreground and the background pixels in the label.

\subsection{Training Details}

To train the model, we use some video sequences from the CDnet 2014 dataset \cite{CDnet2014}.  The same with \cite{bianco2017combination},   the shortest video sequence  from each category is chosen. These sequences are (sequence/category): pedestrians/baseline,  badminton/cameraJitter, canoe/dynamicBackground, parking/intermittentObjectMotion, peopleInShade/shadow, park/thermal, wetSnow/badWeather, tramCrossroad\_1fps/lowFramerate, winterStreet/nightVideos, zoomInZoomOut/PTZ and turbulence3/turbulence.
For each sequence, a subset of frames which contains many foreground objects are selected as the training frames, the frames with  only background or few foreground objects are not used to training the model. Thus we can see that the training sequences (11/53) and the training frames ($\sim$4000/$\sim$160000) are only a small part of the total dataset, which guarantees the generalization power of our model.

To make a fair comparison with other combination strategies such as \cite{CDnet2014, bianco2017combination}, we take the output results from SuBSENSE \cite{subsense}, FTSG \cite{FTSG}, and CwisarDH \cite{cwisardH} algorithms as the benchmark. As illustrated in Fig. \ref{iutis3}, during the training stage, three foreground/background masks produced by these BGS methods are pre-resized to a size of $224\times224\times1$, then concatenated as a 3 channels image and fed into the network.
For the label masks, the label value is given by:
\begin{equation}
y_p =\begin{cases}
1, & \text{if   class($p$) = foreground }      ;\\
0, & \text{otherwise. }
\end{cases},
\label{GT_labels}
\end{equation}
where $p$ denotes the pixels in the label masks.

The proposed network is implemented in TensorFlow \cite{Tensorflow}. We fine-tune the entire network for 50 epochs. The Adam optimization strategy is used for updating  model parameters. 

\section{Experimental results}\label{Experimental results}

\begin{table*}[!h]
\begin{center}
\caption{Complete results evaluated  on the CDnet 2014 dataset}
{\tabcolsep15pt\begin{tabular}{cccccccc}
\toprule
Category&Re&Sp&FPR&FNR&PWC&Pr&FM\\
\midrule

baseline        & 0.9376 & 0.9986 & 0.0014  &0.0624 & 0.4027 &	0.9629 &	0.9497\\
cameraJ         & 0.7337 & 0.9965 & 0.0035  &0.2663 & 1.5542 & 	0.9268 &	0.8035\\
dynamic         & 0.8761 & 0.9997 & 0.0003  &0.1239 & 0.1157 &  0.9386 &	0.9035\\
intermittent    & 0.7125 & 0.9960 & 0.0040  &0.2875 & 3.2127 &	0.8743 &	0.7499\\
shadow          & 0.8860 & 0.9974 & 0.0026  &0.1140 & 0.8182 & 0.9432  &	0.9127\\
thermal         &0.7935  & 0.9970 & 0.0030  &0.2065 & 1.5626 & 0.9462  &	0.8494\\
badWeather      &0.8599  & 0.9996 & 0.0004  &0.1401 & 0.3221 & 0.9662  &	0.9084\\
lowFramerate    &0.7490  & 0.9995 & 0.0005  &0.2510 & 1.0999 & 0.8614  &	0.7808\\
nightVideos     &0.6557  & 0.9949 & 0.0051  &0.3443 & 1.2237 & 0.6708  &	0.6527\\
PTZ             &0.6680  & 0.9989 & 0.0011  &0.3320 & 0.4335 & 0.8338  &	0.7280\\
turbulence      &0.7574  & 0.9998 & 0.0002  &0.2426 & 0.0804 & 0.9417  &	0.8288\\
\midrule
\bf{Overall}    &0.7845  & 0.9980 & 0.0020  &0.2155 & 0.9842 & 0.8969  &    0.8243\\
\bottomrule
\end{tabular}}{}
\label{resultsOfCD2014}
\end{center}
\end{table*}

\begin{table}[!h]
\begin{center}
\caption{Performance comparison of different fusion strategy}
{\tabcolsep8pt\begin{tabular}{cccc}
\toprule
 Strategy                                      & Recall & Precision & F-Measure\\
\midrule
IUTIS-3 \cite{bianco2017combination}                         &      \bf{0.7896}&	      0.7951     &          0.7694             \\
MV          \cite{CDnet2014}                                 &         0.7851  &	      0.8094     &          0.7745             \\
CNN-SFC (\textcolor[rgb]{1.00,0.00,0.00}{our})                &         0.7845  &	   \bf{0.8969}   &       \bf{0.8243}             \\
\bottomrule
\end{tabular}}
\label{fusioncomparsion}
\end{center}
\end{table}

\begin{table*}[ht]
\begin{center}
\caption{Comparison of the results on the CDnet 2014 dataset by different BGS algorithms}
{\tabcolsep11pt\begin{tabular}{ccccccccc}
\toprule
 Method     & Ranking          & Re   & Sp  & FPR   & FNR  & PWC  & Pr   & FM \\
\midrule
\textcolor{blue}{CNN-SFC} \textcolor{red}{ (our)}
& 7.27                                   & 0.7709              & 0.9979           & 0.0021         & 0.2291         & 1.0409           & 0.8856             &	0.8088\\ 
\textcolor{blue}{IUTIS-5 \cite{bianco2017combination}}
& 8.27                                   & 0.7849              & 0.9948           & 0.0052         & 0.2151         & 1.1986           & 0.8087             &	0.7717\\ 
\textcolor{blue}{IUTIS-3 \cite{bianco2017combination}}
&12.27                                    & 0.7779              & 0.9940           & 0.0060         & 0.2221         & 1.2985           & 0.7875             &	0.7551\\ 
\textcolor{blue}{SuBSENSE \cite{subsense}}
&15.55                                   & 0.8124              & 0.9904           & 0.0096         & 0.1876         & 1.6780           & 0.7509             &	0.7408\\ 
\textcolor{blue}{FTSG \cite{FTSG}}
&15.55                                   & 0.7657              & 0.9922           & 0.0078         & 0.2343         & 1.3763           & 0.7696             &	0.7283\\ 
\textcolor{blue}{CwisarDH \cite{cwisardH}}
&22.18                                   & 0.6608              & 0.9948           & 0.0052         & 0.3392         & 1.5273           & 0.7725             &	0.6812\\ 
\textcolor{blue}{KDE \cite{KDE}}
&33.27                                   & 0.7375              & 0.9519           & 0.0481         & 0.2625         & 5.6262           & 0.5811             &	0.5688\\ 
\textcolor{blue}{GMM \cite{GMM}}
&36.91                                   & 0.6604              & 0.9725           & 0.0275         & 0.3396         & 3.9953           & 0.5975             &	0.5566\\ 
\bottomrule
\end{tabular}}
\label{compareWithOthers}
\end{center}
\end{table*}

\subsection{Dataset and Evaluation Metrics}\label{paramtersN}

We based our experiments on the CDnet 2014 dataset \cite{CDnet2014}. CDnet 2014 dataset  contains   53   real scene video sequences with nearly 160 000 frames.  These sequences are grouped into 11 categories corresponding different challenging conditions. They are   baseline,  camera jitter, dynamic background,  intermittent object motion, shadow, thermal, bad weather, low framerate, night videos, pan-tilt-zoom, and turbulence.
Accurate human expert constructed ground truths are available for all sequences and seven   metrics have been defined in \cite{CDnet2014} to compare the performance of different algorithms: 
\begin{itemize}

\item Recall ({Re}) = $\frac{TP}{TP+FN}$

\item Specificity ({Sp}) = $\frac{TN}{TN+FP}$

\item False positive rate ({FPR}) = $\frac{FP}{FP+TN}$

\item False negative rate ({FNR}) = $\frac{FN}{TP+FN}$

\item Percentage of wrong classifications ({PWC}) = $100 \cdot\frac{FN+FP}{TP+FN+FP+TN}$

\item Precision ({Pr}) = $\frac{TP}{TP+FP}$

\item F-Measure ({FM}) = $2\cdot\frac{Re\cdot Pr}{Re+Pr}$
\end{itemize}
where  \textit{TP} is true positives, \textit{TN} is true negatives, \textit{FN} is false negatives, and \textit{FP} is false positives.
For {Re, Sp, Pr} and {FM} metrics, high score values indicate better performance, while for {PWC, FNR} and {FPR}, the smaller the better.
Generally speaking, a BGS algorithm is considered good if it gets high recall scores  without sacrificing precision. So, the FM metric is a good indicator of the overall performance.
As shown in \cite{CDnet2014}, most state-of-the-art BGS methods  usually obtain higher FM scores than other worse performing methods.

\subsection{Performance Evaluation}

\textit{Quantitative Evaluation}: Firstly, in Table \ref{resultsOfCD2014}, we  present the  evaluation results of the  proposed method using the evaluation tool provided by the CDnet 2014 dataset \cite{CDnet2014}. Seven metrics scores, as well as the overall performance for each sequence  are presented.  As we stated earlier, we use  the  SuBSENSE \cite{subsense}, FTSG \cite{FTSG}, and CwisarDH \cite{cwisardH} algorithms as the benchmark. According to the reported results\footnote{\color{blue}{{www.changedetection.net}}}, they achieved an initial FM metric score of 0.7453, 0.7427 and 0.7010  respectively on the dataset.  However, through the proposed fusion strategy, we achieve an FM score of 0.8243, which is a significant improvement (11\%) compared with the best 0.7453(SuBSENSE).

Secondly, to demonstrate our key contribution, the proposed fusion strategy (CNN-SFC) is preferable to others \cite{CDnet2014}, \cite{bianco2017combination}. In Table \ref{fusioncomparsion}, we give the performance comparison   results of different fusion strategies applied on the  SuBSENSE \cite{subsense}, FTSG \cite{FTSG}, and CwisarDH \cite{cwisardH} results.
We can see that our combination strategy achieves a much higher FM score than the majority vote and genetic programming strategies, this is mainly benefited from the huge improvement of the precision metric.  For the recall metric, all fusion strategies are almost the same since recall measures.
The recall is the ratio of the number of foreground pixels correctly identified by the BGS algorithm to the number of foreground pixels in ground truth. This is mainly determined by the original BGS algorithms, so, all fusion strategies  almost have the same score. However, the precision is defined as the ratio of the number of foreground pixels correctly identified by the algorithm to the number of foreground pixels detected by the algorithm. From Fig. \ref{iutis3}, we can see that after the training process, the neural network learn to leverage the characteristics of different BGS algorithms. In the final output result, many false positive and false negative pixels are removed, so that the precision of CNN-SFC is much higher than MV and IUTIS.

%
%
%
%

\begin{figure*}[htbp]
\centering
\includegraphics[width=1.0\linewidth]{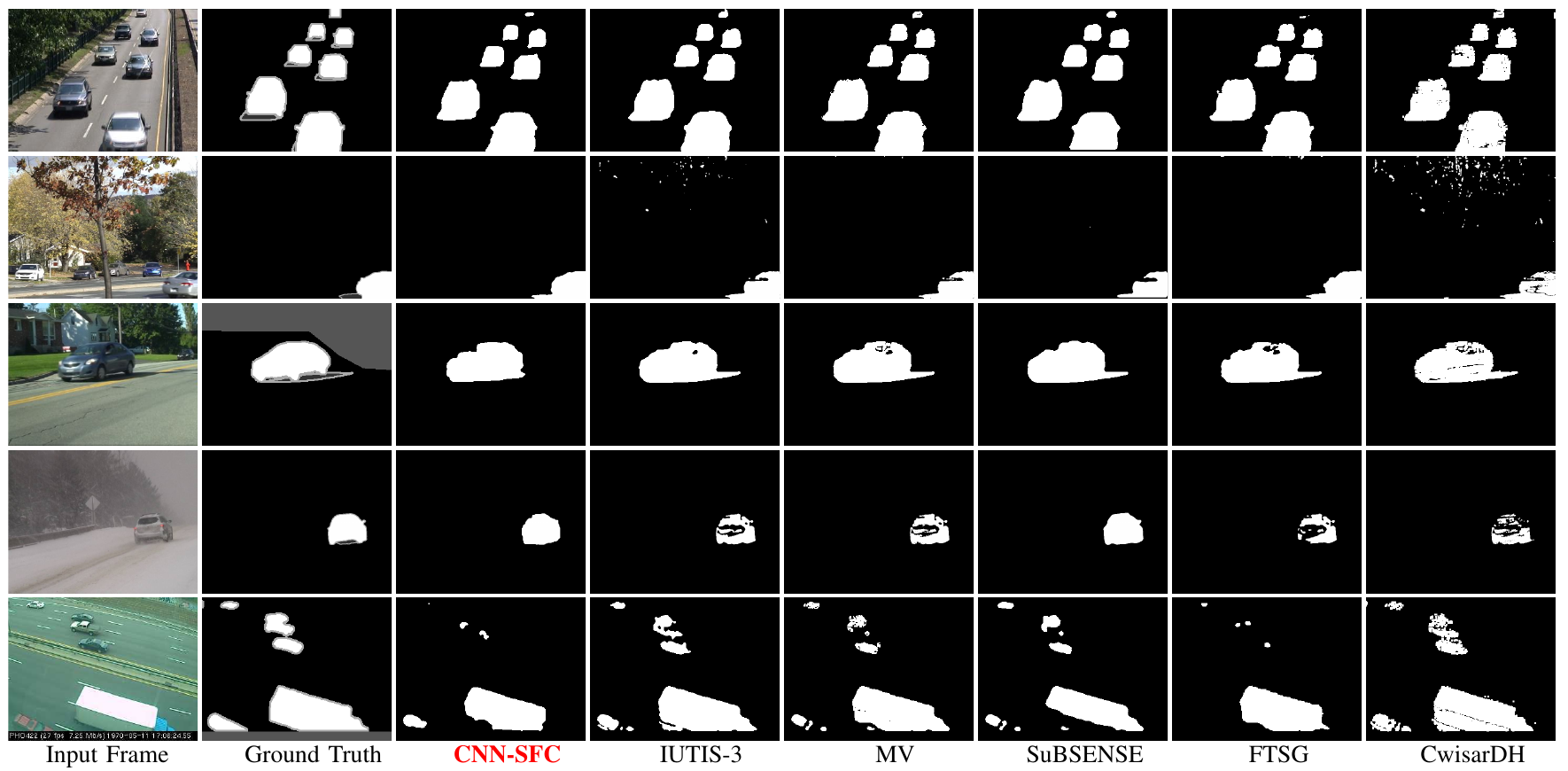}   
\caption{Qualitative performance comparison for various sequences (from top to bottom: \textit{highway, fall, bungalows, snowFall}  and \textit{turnpike\_0\_5fps} ).  The first column to the last column: input frame, ground truth, our  result, IUTIS-3 \cite{bianco2017combination},  SubSENSE \cite{subsense}, FTSG \cite{FTSG} and  CwisarDH \cite{cwisardH} detection results.}
\label{QualitativeComparisonFigure}
\end{figure*}

Finally,   we submitted our results to the website\footnote{\color{blue}{{http://jacarini.dinf.usherbrooke.ca/results2014/529/}}} and made a comparison with the  state-of-the-art BGS methods:  
{IUTIS-5 \cite{bianco2017combination}},   {IUTIS-3 \cite{bianco2017combination}},  {SuBSENSE \cite{subsense}},  {FTSG \cite{FTSG}},  {CwisarDH \cite{cwisardH}},  {KDE \cite{KDE}}, and {GMM \cite{GMM}}.
The results are shown in Table \ref{compareWithOthers}, here, we mainly make a comparison between the unsupervised BGS algorithms. For the supervised BGS methods, ground truths selected from each sequence  are used to train their models. And the trained model is difficult to generalize to other sequences that have not been seen(trained) before. Thus, these methods should not be compared directly with the other unsupervised methods.  We can see that the proposed method ranks the first among all unsupervised BGS algorithms.

%

\textit{Qualitative evaluation}: To make a  visual comparison of these BGS methods, some typical segmentation results  under different scenarios are shown in Fig. \ref{QualitativeComparisonFigure}. The following frames are selected: the 831th frame from the \textit{highway} sequence of baseline category , the 1545th frame from the \textit{fall} sequence of the dynamic background category,  the 1346th frame from the \textit{bungalows} sequence of the shadow category, the 2816th frame from the \textit{snowFall} sequence of the bad weather category and the 996th frame from the \textit{turnpike\_0\_5fps} sequence of the low framerate category.
In Fig. \ref{QualitativeComparisonFigure}, the first column displays the input frames and the second column shows  the  corresponding ground truth. From the third column to the eighth column, the foreground objects detection results of the following method are showed: our method (CNN-SFC),  IUTIS-3, MV, SuBSENSE,  FTSG, and CwisarDH.
Visually, we can see that our results look much better than   other fusion  results and the benchmark BGS results. This is confirmed with the quantitative evaluation results.

\section{CONCLUSION}\label{CONCLUSION}

In this paper, we propose an encoder-decoder fully convolutional neural network for combining the foreground/backgroud masks  from different state-of-the-art background subtraction algorithms. Through a training process, the neural network learns to leverage the characteristics of different BGS algorithms, which produces a more precise foreground detection result. 
Experiments evaluated on the CDnet 2014 dataset show that the proposed combination strategy is much more efficient than  the majority vote and genetic programming based fusion strategies. The proposed method is currently ranked the first among all unsupervised BGS algorithms.

\section*{Acknowledgment}
This work was supported by  the National Nature Science Foundation of China under Grant No.61401425.
We gratefully acknowledge the support of NVIDIA Corporation with the donation of the Titan Xp GPU used for this research.

{
\footnotesize

}

\end{document}